# The Model Is Not the Product: A Dual-Pillar Architecture for Local-First Psychological Coaching

*Combining Foundation Language Models With Sophisticated Active Learning Memory Systems for Considerable On-Device Performance*

Alexander R. Mihalcea

May 2026

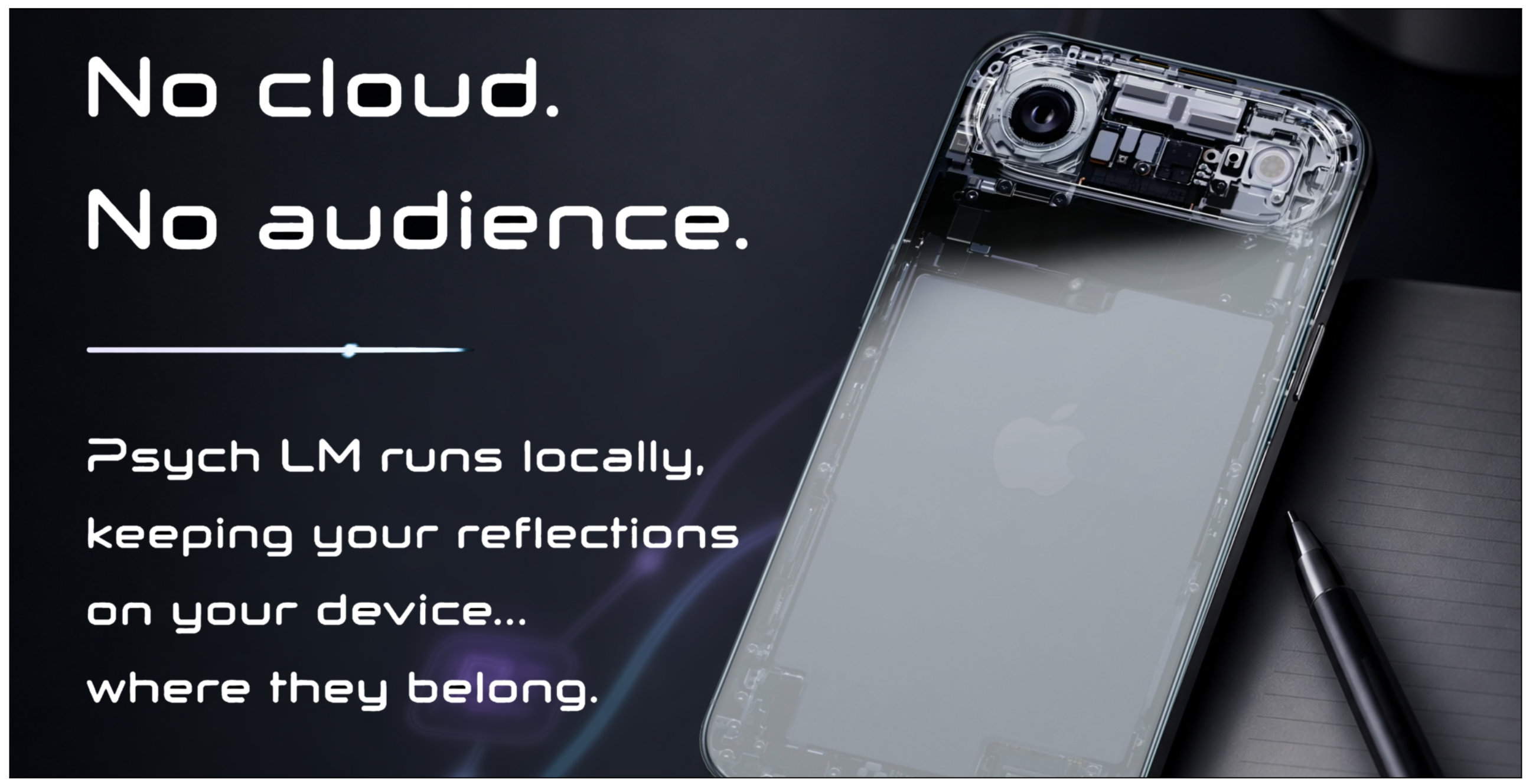

**Abstract-** Existing language model applications struggle to meet the demand for emotionally oriented support, primarily due to their inability to maintain deep, persistent context across sessions (Lawrence et al., 2024; Li et al., 2023). This report introduces Psych LM, an iOS application that validates the thesis that for such applications, the surrounding architecture is paramount (Boit & Patil, 2025; Lewis et al., 2020; Packer et al., 2023). Psych LM runs a local, on-device language model within a purpose-built, local-first runtime designed for behavioral and life-coaching applications. The system achieves the practical effect of a near-infinite context window through an automated, user-inspectable memory corpus that converts conversations into structured "memory cards" (facts, goals, events) and dynamically injects them into the prompt via semantic and vector search (Lewis et al., 2020; Packer et al., 2023; Reimers & Gurevych, 2019). As such, the system could be defined as an active-learning, retrieval-augmented generative, on-device architecture. This architecture delivers four primary contributions: a local-first design where privacy is a core property; a detailed description of the memory corpus for persistent context of key user information; a deterministic orchestration layer that provides a stable behavioral spine independent of the model's internal state; and a benchmark framework focused on evaluating the integrated system's reliability under realistic operating conditions. The R&D process confirms that complex, context-aware interaction can be reliably achieved under the strict constraints of a mobile environment by prioritizing architectural control and resource management over simple model size.

## 1. Introduction

In recent years, the power and versatility of language models have expanded rapidly across nearly every domain, from the arts to the hard sciences. One particularly notable use case is in emotionally oriented applications, including forms of pseudo-therapeutic interaction (Lawrence et al., 2024; Li et al., 2023). This trend has drawn significant criticism from professionals in the field (Boit & Patil, 2025; Lawrence et al., 2024). However, the frequency with which these systems are used in this way suggests the presence of a real and unmet need—one that the current professional infrastructure is unable to fully address (Lawrence et al., 2024; Li et al., 2023).

Several factors likely contribute to the appeal of language model–based systems in this context. These include anonymity, reduced fear of judgment, lower cost, and increased convenience. While leading providers such as OpenAI, Anthropic, and Google offer highly sophisticated models, this work emphasizes a central premise: the model itself represents only a small portion of the overall system. The surrounding architecture is often underdeveloped and underappreciated in existing approaches.

### *1.1. Motivation*

For a language model–based system to provide meaningful support in domains such as behavioral or psychological coaching, it must satisfy several key criteria. The most critical requirement is persistent context across sessions (Packer et al., 2023; Park et al., 2023). True usability depends on the system's ability not only to retain general user information, but to store and reason over specific events, their temporal structure, and their relationships to broader themes in the user's life.

This includes continuity from prior sessions, including where a previous interaction concluded. Most existing online systems struggle with this requirement (Packer et al., 2023). While some incorporate limited memory functionality, these mechanisms are not designed for deep, structured, longitudinal context tracking.

### *1.2. Core Thesis*

This report introduces Psych LM, an iOS application that runs one of several local, on-device language models within a purpose-built architecture designed for behavioral, psychological, and general life-coaching applications. The system distinguishes itself through several key features. Most notably, all functionality is executed entirely on-device, including model inference, response generation, memory creation, storage, and retrieval. Once the required model files are installed, all operations occur locally on the user's device (Liu et al., 2024; Xu et al., 2024).

Consistent with the core thesis, Psych LM is designed such that the language model itself is only one component of a larger system. A second core pillar is the memory corpus. The memory corpus functions as a structured repository in which the system automatically generates "memory cards" during user interaction. These cards represent key events, topics, or other salient elements of the user's life and are categorized accordingly (Packer et al., 2023; Park et al., 2023). During subsequent interactions, these memory cards can be retrieved using a combination of semantic and vector-based search (Douze et al., 2024; Karpukhin et al., 2020; Reimers & Gurevych, 2019), as illustrated in Fig. 1.

The interaction between the language model and the memory corpus is the primary factor enabling the system's effectiveness. By storing specific events, their temporal anchors, and their relationships, and by retrieving them dynamically, the system creates the practical effect of a near-infinite context window despite operating with a relatively small on-device model (Packer et al., 2023). This capability is central to the system's functionality.

### *1.3. Contributions*

This paper makes four primary contributions.

First, it presents a local-first runtime architecture for an on-device conversational system. In this implementation, Psych LM is designed around local inference and local persistence, meaning that dependence on cloud infrastructure is eliminated at the architectural level rather than treated as a secondary privacy feature. This positions privacy not as an external assurance layered onto the system, but as a structural property of the runtime itself (Iwaya et al., 2023; Kleppmann et al., 2019).

Second, it describes an automated memory corpus intended to support persistent user context. Conversations are converted into structured memory cards representing facts, preferences, goals, plans, values, boundaries, and risk-relevant flags. These memories are then indexed, retrieved, injected into prompts, and made available for user inspection and correction (Lewis et al., 2020; Packer et al., 2023), thereby allowing persistent personalization to remain both operationally useful and procedurally inspectable.

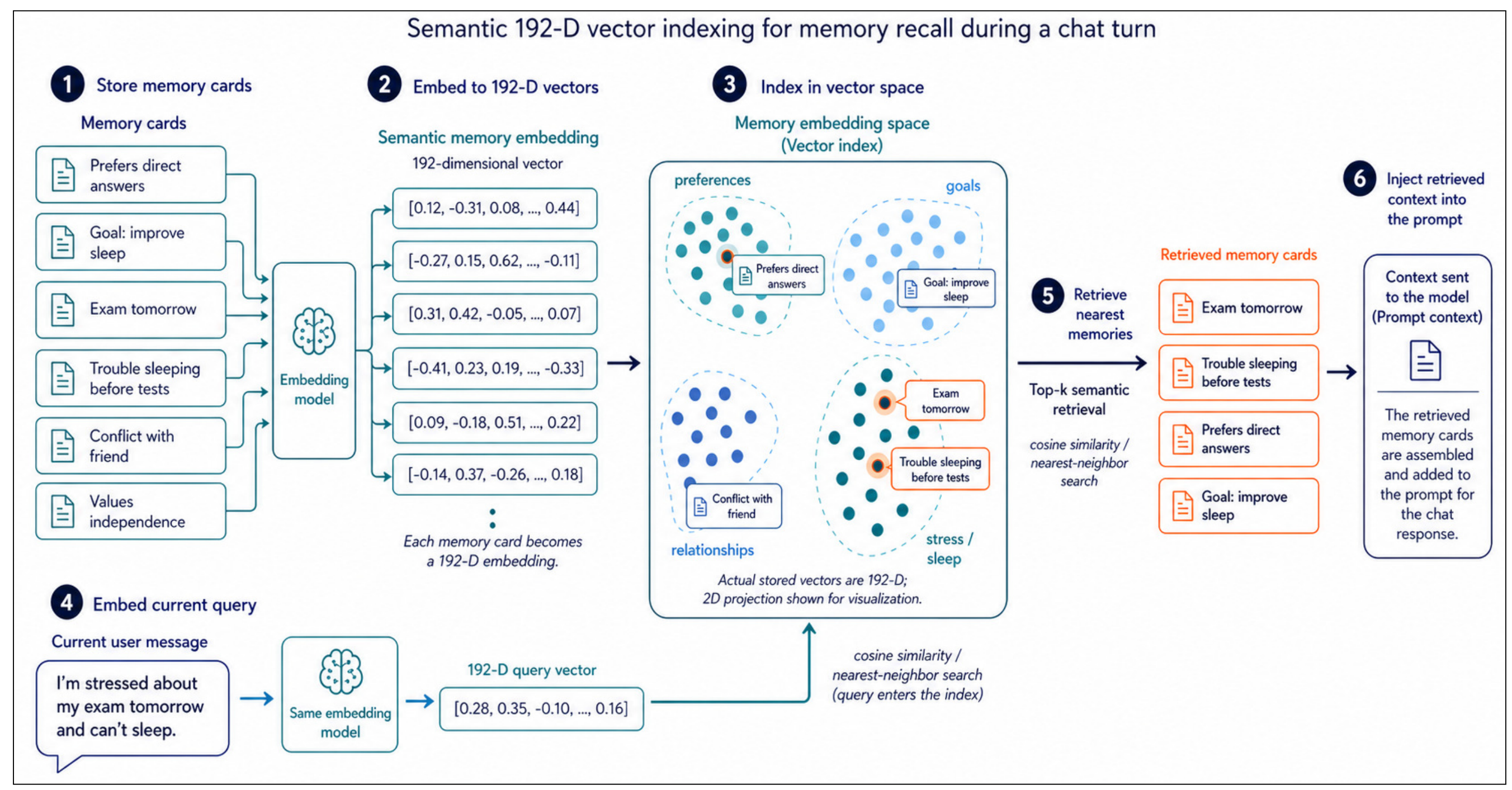

Fig. 1. Illustration of the memory corpus structure used to support semantic and vector-based memory retrieval.

Third, it describes a deterministic orchestration layer surrounding the generative model (Boit & Patil, 2025; White et al., 2023). Intake progression, prompt construction, context assembly, memory retrieval, output parsing, safety handling, and persistence are controlled by the application rather than delegated entirely to the model. This distinction is important because it gives the system a stable behavioral spine, reducing the extent to which core application behavior depends on the model's variable and sometimes unreliable internal state.

Fourth, it proposes a benchmark framework for evaluating the system as a deployed pipeline rather than as an isolated model. The evaluation covers runtime performance, memory use, personality control, and safety-boundary behavior (Liang et al., 2023). In this context, the benchmarks are intended to measure whether the integrated system behaves reliably under realistic operating conditions, rather than merely whether the underlying model appears capable in abstraction.

## 2. Persistent User and Temporal Context

The second design goal is persistence, with temporal understanding treated as a first-order part of that persistence rather than as a secondary prompt detail. Psych LM is not intended to behave like a stateless chat window that begins each interaction without meaningful continuity (Packer et al., 2023; Park et al., 2023). Instead, it maintains sessions, user profile data, goals, preferences, long-term memory cards, session digests, and chronology-relevant context that can be retrieved in later interactions.

This goal is especially important given the nature of the underlying model. The local language model should not be treated as intrinsically aware of the present moment, the user's prior timeline, or the temporal relationship between past and current events. For that reason, the application explicitly provides temporal grounding at inference time, including the current date and time as a concrete string rather than relying on the model to infer or hallucinate temporal context. This makes time an application-supplied constraint, not an assumed property of the model.

Persistent context therefore serves three related functions. First, it improves continuity by allowing the system to refer back to prior goals, plans, preferences, and user-provided information without requiring the user to reconstruct their full situation in every session. Second, it creates a substrate for personalization that exists outside the model weights. The model does not need to be fine-tuned on the user in order to behave in a more personally relevant manner; rather, the application can maintain a local memory corpus and inject relevant context at inference time (Lewis et al., 2020; Packer et al., 2023). Third, it allows the system to reason more coherently about temporal relationships, such as whether a goal was created recently, whether a stressor is recurring, whether a plan is overdue, or whether a user's current statement should be interpreted against an earlier session.

This design also makes certain failure modes more inspectable. If the assistant uses an incorrect fact, misorders events, or responds as though it does not know the current date, the error can be traced to a more specific component, such as memory extraction, chronology construction, retrieval, prompt assembly, explicit date-time injection, or output generation (Boit & Patil, 2025; Packer et al., 2023). Without an explicit memory and temporal-grounding layer, similar failures are more likely to collapse into vague "model behavior," which is substantially harder to debug because the causal pathway between stored context, current time, and generated output remains under-specified.

## 3. R&D Insights

The development process for Psych LM produced several practical findings that were not fully apparent from the initial system concept. These findings should not be interpreted as evidence that the current architecture is the final, complete, or ideal way to build a local-first psychological reflection system. Rather, they describe the constraints, failures, and design adjustments that emerged while attempting to make a memory-augmented, on-device conversational system behave reliably in ordinary use.

### *3.1. Prompt Payload as a Primary Design Surface*

The first major finding was that prompt payload assembly mattered substantially more than initially expected. The relevant problem was not merely prompt wording, but the composition of the entire prompt packet: personality instructions, coaching approach, retrieved memory, user profile data, temporal context, recent messages, safety constraints, and task-specific directives (Sahoo et al., 2024; White et al., 2023). If the payload became too large, the model became slower and more vulnerable to drift; if it became too small, responses lost the context needed to feel continuous or personally relevant. This made prompt assembly a core runtime concern rather than a secondary implementation detail.

In response, the prompt pipeline became increasingly compartmentalized. Personality, therapeutic approach, retrieved context, temporal grounding, recent history, and task-specific instructions were treated as dynamic prompt components rather than as a single static system message. This allowed the application to make better use of a limited context budget, especially when running smaller local models. In this respect, the prompt became less like a fixed instruction block and more like a resource allocation problem, where each section had to justify its inclusion against latency, drift risk, and context-window pressure (Sahoo et al., 2024; Xu et al., 2024).

### *3.2. Mobile Performance Outside the Model Runtime*

A related finding was that mobile performance bottlenecks did not always originate inside the model runtime itself. One particularly concrete example was the live rendered glow animation, which was eventually found to reduce tokens per second by roughly an order of magnitude. This was useful because it revealed that local inference performance is not only a property of model size, quantization, or sampling configuration (Alizadeh et al., 2023; Xu et al., 2024). It is also affected by the surrounding application environment, including UI rendering behavior, thermal load, and competing work on the device. The magnitude of this GUI-related performance effect was unexpected, and the issue took several weeks to discover and resolve.

### *3.3. Memory Generation Before Memory Retrieval*

Memory development produced a separate and more difficult lesson. The hardest part was not initially retrieval, but the generation of useful memory cards in the first place (Packer et al., 2023; Park et al., 2023). Many candidate memories were low-value, incorrect, repetitive, derived from assistant statements rather than user-provided facts, or simply hallucinated. Conversely, some highly information-rich sessions produced no useful memory cards at all. This made it impracticable to treat memory creation as a simple extract-and-store operation. The system required review, filtering, validation, and app-mediated persistence because the upstream memory-generation process was itself noisy and unreliable.

This finding also clarified why memory needed to remain inspectable. A hidden memory layer would have made these failures substantially harder to diagnose, because incorrect personalization would appear to be vague model behavior rather than a traceable error in extraction, retrieval, or prompt assembly. By exposing memory cards as a user-facing state, the system makes it possible to identify whether a poor response came from an incorrect card, a missed memory, an irrelevant retrieval, or a generative failure (Kleppmann et al., 2019; Packer et al., 2023). This does not eliminate memory errors, but it makes them more procedurally visible.

### *3.4. Deterministic Control Around Reasoning Models*

Deterministic orchestration became more important as different model families were tested. Reasoning-oriented models, particularly the now-deprecated Qwen-based reasoning paths, did not reliably obey requested limits on chain-of-thought length. In practice, ordinary prompt instruction was insufficient to control the reasoning phase, which required a more direct output-boundary intervention: forcing the end-thinking marker or answer boundary into the generation path so that the system could terminate the reasoning phase and recover usable assistant output. This illustrates a broader principle: when model prompting is insufficient for a task, control often has to move from prompting into runtime or orchestration logic (Boit & Patil, 2025; White et al., 2023).

### *3.5. Model Lineup and Selection Criteria*

Model selection evolved in a similar way. Apple's Foundation Models were initially attractive because of their expected speed and system integration, but their refusal behavior made them unsuitable for nearly any purpose within the app. Gemma became the preferred small-model path because it offered a comparatively strong balance of responsiveness, personality adherence, and general compliance, although it still required careful prompting. Llama served for much of development as a middle-tier option, but licensing and naming constraints introduced product friction, and its practical advantage narrowed as Gemma models became more effective under the revised prompt pipeline. Qwen initially occupied the large reasoning-model role, but reasoning-control issues and the broader capabilities of the later Gemma 4 lineup eventually favored a more unified Gemma-centered model strategy.

The broader lesson was that model selection could not be reduced to raw benchmark quality (Liang et al., 2023; Liu et al., 2024). For this product, model usefulness depended on behavioral reliability, prompt obedience, licensing constraints, runtime fit, future extensibility, and whether the model could be integrated without distorting the surrounding application architecture.

### *3.6. Safety Layer Deprecation and Boundary Behavior*

The safety work produced a more qualified result than expected. An earlier auxiliary safety system, designed to run above the model, misfired often enough that it became impracticable and was eventually deprecated. This did not mean that safety boundaries were unnecessary. Rather, it suggested that an overly sensitive external safety layer can degrade ordinary reflective use, especially in a domain where benign conversation may still contain emotionally charged language (Boit & Patil, 2025; Lawrence et al., 2024).

Within reasonable test environments, the local models were generally capable of avoiding harmful advice, though they were less reliable at consistently detecting acute crisis language and producing the desired response. The resulting lesson was not that safety was solved, but that safety control must be balanced against over-triggering, usability loss, and the limits of small-model crisis detection (Lawrence et al., 2024). Nevertheless, all models' pretrained safety systems proved generally robust enough for the app's intended use case.

### *3.7. Intake, Check-Ins, and User Tolerance*

The intake system also changed substantially during development. The original goal was to create a model-led intake flow that more closely resembled a human conversational onboarding process. However, this approach proved too tedious in practice, even after attempts at optimization. The more important product lesson was that conversational naturalism is not automatically better than structured interaction. A flow can be technically interesting and still fail if it delays the user's first meaningful experience with the product.

This led to a stronger emphasis on check-ins rather than a long model-led intake. The check-in design preserved the useful part of the original idea: users could answer in free-response form, while the language model performed the mapping between natural language and structured scoring criteria. This made the model useful as an interpreter of user language without requiring the entire onboarding process to become a simulated interview. In practical terms, the model was more valuable when embedded inside a bounded structured workflow than when allowed to control the interaction sequence outright (Boit & Patil, 2025). This also led to a modest innovation: although the questionnaires in the app are surrogates of their true clinical counterparts and are not meant for clinical use, the ability to answer them in natural language—rather than selecting one of a few prewritten options—proved promising in testing as a potential future avenue to explore more seriously.

### *3.8. Beta Feedback and Feature Discipline*

Beta testing further shifted the product direction from feature expansion toward feature discipline. Users were often most impressed by the system's ability to remember basic information across chats, because this made the assistant feel continuous rather than stateless. UI polish also received substantial positive feedback, suggesting that perceived quality depended not only on model capability but also on visual coherence, responsiveness, and general usability (Li et al., 2023). As a result, development shifted away from continuously adding new features and toward pruning unused surfaces while polishing the smaller set of features that directly affected user satisfaction.

### *3.9. Local-First Design as a Product Strategy*

Local-first design was similarly clarified through development. The decision to run on device was not only a privacy decision, but also an access and distribution decision. From the beginning, the goal was to offer a fully functional free version of the app, without chat limits, advertising, or dependence on metered hosted inference. On-device execution made that product model more viable, because it aligned privacy, offline availability, and cost control under the same architectural decision (Kleppmann et al., 2019; Xu et al., 2024).

However, local inference also imposed substantial technical costs. Latency, battery use, thermal behavior, app size, model-download burden, and memory pressure all became first-order design constraints (Alizadeh et al., 2023; Liu et al., 2024; Xu et al., 2024). These constraints interacted with prompt size, model selection, UI rendering, and retrieval strategy, meaning that the system could not be optimized one layer at a time in isolation. The practical R&D conclusion is that local-first conversational AI is not merely a smaller version of cloud inference. It requires the entire application to behave as a resource-managed pipeline.

### *3.10. Bounded Interpretation*

Taken together, these findings support a narrower and more defensible interpretation of the project. Psych LM does not demonstrate that local models, memory systems, or deterministic orchestration are sufficient to solve psychological support. It demonstrates that, for this class of application, the surrounding system architecture is at least as important as the model itself (Boit & Patil, 2025; Lewis et al., 2020; Packer et al., 2023). The model generates text, but the product behavior emerges from prompt assembly, memory validation, retrieval, runtime control, UI performance, safety boundaries, and user-facing inspectability. The R&D process therefore supports the central thesis of the paper while also limiting it: the current system is a practical implementation pathway, not a final answer.

## 4. System Architecture

Psych LM is organized as a set of cooperating application subsystems rather than as a conventional chat interface wrapped around a language model. At the implementation level, the system is best understood as a locally running, active-learning, retrieval-augmented, safety-gated yet adaptive architecture (Boit & Patil, 2025; Lewis et al., 2020). It generates text and, in some cases, structured sidecar outputs, but the application determines what context is supplied, which outputs are accepted, how state is validated, and what information becomes durable (Boit & Patil, 2025; White et al., 2023).

### *4.1. Top-Level Architecture*

At the highest level, Psych LM is organized as a pipeline of cooperating layers: presentation, persistence, orchestration, prompt construction, memory corpus management, model runtime, and output control. These layers are functionally separable, but the user-visible response depends on their coordination rather than on the language model alone (Boit & Patil, 2025; Liang et al., 2023).

The presentation layer exposes chats, memory, goals, settings, model controls, and voice surfaces, while the persistence layer stores the local data needed for longitudinal behavior. The orchestration layer coordinates each turn by resolving session state, retrieving context, constructing the prompt, invoking the local runtime, parsing the result, and committing validated state changes.

Prompt construction converts application state into model-facing context, including instructions, profile data, temporal grounding, personality settings, recent dialogue, retrieved memory, and task-specific directives. The memory and corpus layer supplies durable user context outside the model weights (Lewis et al., 2020; Packer et al., 2023), allowing the system to selectively reintroduce relevant information without treating the model itself as the source of long-term knowledge.

The runtime layer executes local inference through the selected model path, while output-control logic constrains what becomes visible or durable. Together, these layers make Psych LM a locally persisted, memory-augmented, pipeline-controlled system rather than a model with a chat interface attached to it.

### *4.2. Chat Turn Pipeline*

A standard conversation turn is controlled by the application rather than passed directly from user input to model output. The user message is first validated and persisted, after which the system resolves the active profile, session state, settings, recent messages, and relevant long-term context.

The memory subsystem retrieves a limited set of prompt-relevant cards or corpus entries, which are treated as selected contextual evidence rather than as a complete user biography (Lewis et al., 2020; Packer et al., 2023). This search was critical because it needed to find the most relevant few cards out of potentially hundreds while introducing almost no latency on device. To achieve this, a combination of semantic search and model-based embeddings is used to balance relevance and retrieval speed (Karpukhin et al., 2020; Reimers & Gurevych, 2019). PromptBuilder assembles these materials with instructions, temporal grounding, personality settings, recent dialogue, and the current user message before passing the resulting payload to the selected runtime.

After inference, the model output is parsed, cleaned, checked against scope constraints, and only then persisted. Any memory updates are treated as application-mediated state changes rather than direct model writes (Kleppmann et al., 2019; Packer et al., 2023). Thus, the turn pipeline is best summarized as:

user input → state resolution → memory retrieval → prompt construction → inference → output control → persistence

The central point is that inference is only one bounded stage inside a larger orchestration loop.

### *4.3. Runtime and Model Management*

Psych LM uses a runtime abstraction so that model execution can change without requiring the rest of the application to be rewritten. The primary local path is built around GGUF inference, with model metadata, context limits, output targets, and generation parameters managed through application-level configuration rather than scattered assumptions.

This matters because local models operate under strict context, memory, latency, battery, and thermal constraints (Alizadeh et al., 2023; Liu et al., 2024; Xu et al., 2024). Runtime management therefore cannot be separated from prompt construction or memory retrieval. It is part of the same resource-constrained pipeline that determines whether the system remains usable on device. This is further complicated by the nature of iOS, which limits developers from accessing real-time memory-strain information for valid security reasons while also automatically terminating processes that exceed its memory limits. This puts the runtime in a state of ambiguity, requiring careful runtime controls and numerous iterations over several weeks to reach a relatively stable configuration.

## 5. Memory Corpus

The memory corpus is the primary mechanism by which Psych LM maintains continuity across sessions (Packer et al., 2023; Park et al., 2023). It is also the most important distinction between the system and a stateless chat interface.

### *5.1. Memory Card Design*

Psych LM represents durable user context as structured memory cards. Each card captures a small piece of information that may be useful in later conversations. The goal is not to store transcripts verbatim. The goal is to extract key insights (Packer et al., 2023; Park et al., 2023).

Memory cards can represent facts, preferences, goals, plans, values, boundaries, and events. A fact might describe stable user information. A preference might describe how the user wants the assistant to communicate. A goal might capture an active behavioral target (not to be confused with the app's distinct "Goal" feature). A plan might describe a concrete next step. A value might capture a recurring motivation. A boundary might define what the assistant should avoid or how it should frame certain topics. An event may describe a meaningful current or historical occurrence.

This schema is intentionally practical. It does not attempt to model the full complexity of a person, which would be both impossible and ethically dubious. Instead, it creates a compact working memory that can be retrieved when relevant and ignored when not. This allows the system to make the model feel as though it has a near-infinite context window—at least when it functions properly (Packer et al., 2023).

### *5.2. Extraction and Retrieval*

Memory creation occurs after a chat is ended or during the structured intake. When a conversation contains durable information, the system can extract candidate memory cards and store them in the local corpus after first showing them to the user and allowing for manual editing of key details for accuracy. Memory extraction is model-based (Park et al., 2023).

Retrieval occurs before inference. For each user turn, the system forms a retrieval query from the current message and recent context, then selects potentially relevant memory cards. These cards are injected into the prompt so the model can respond with awareness of prior user context. Memory-card lookup is essential at this step, because the feature depends on finding relevant cards while doing so rapidly enough to minimize latency. A mix of semantic and vector search is used to accomplish this (Douze et al., 2024; Karpukhin et al., 2020; Reimers & Gurevych, 2019).

This design separates storage from activation (Lewis et al., 2020; Packer et al., 2023). The system may store many dozens of memory cards, but only a limited subset should be included in a given prompt. Over-retrieval can be as harmful as under-retrieval (Packer et al., 2023). Too little memory produces generic responses. Too much memory bloats the context window and increases the risk that irrelevant facts distort the response. The retrieval layer is therefore a bottleneck, not an accessory.

### *5.3. User Control*

User control is required because memory systems can be wrong. They can store stale information, overgeneralize from a single message, duplicate existing cards, or preserve facts the user no longer wants represented. Psych LM therefore treats memory as editable user-facing state rather than hidden model residue (Iwaya et al., 2023; Kleppmann et al., 2019).

The system includes mechanisms for memory inspection, correction, visualization, backup, and restoration. This supports transparency and gives the user a way to correct the assistant's long-term context. It also makes the system easier to debug. If a response is poorly personalized, the issue may be visible in the memory corpus rather than buried inside an opaque model behavior.

Memory ownership is also part of the local-first design. A persistent assistant should not merely remember the user. It should let the user see what is remembered, change it, and preserve or remove it as needed (Kleppmann et al., 2019). Editing can occur both after initial card generation and whenever the user wishes within the Memory Notebook app sheet. Memory cards can also be manually created by the user and are automatically given the "pinned" tag, meaning they are prioritized during retrieval scoring.

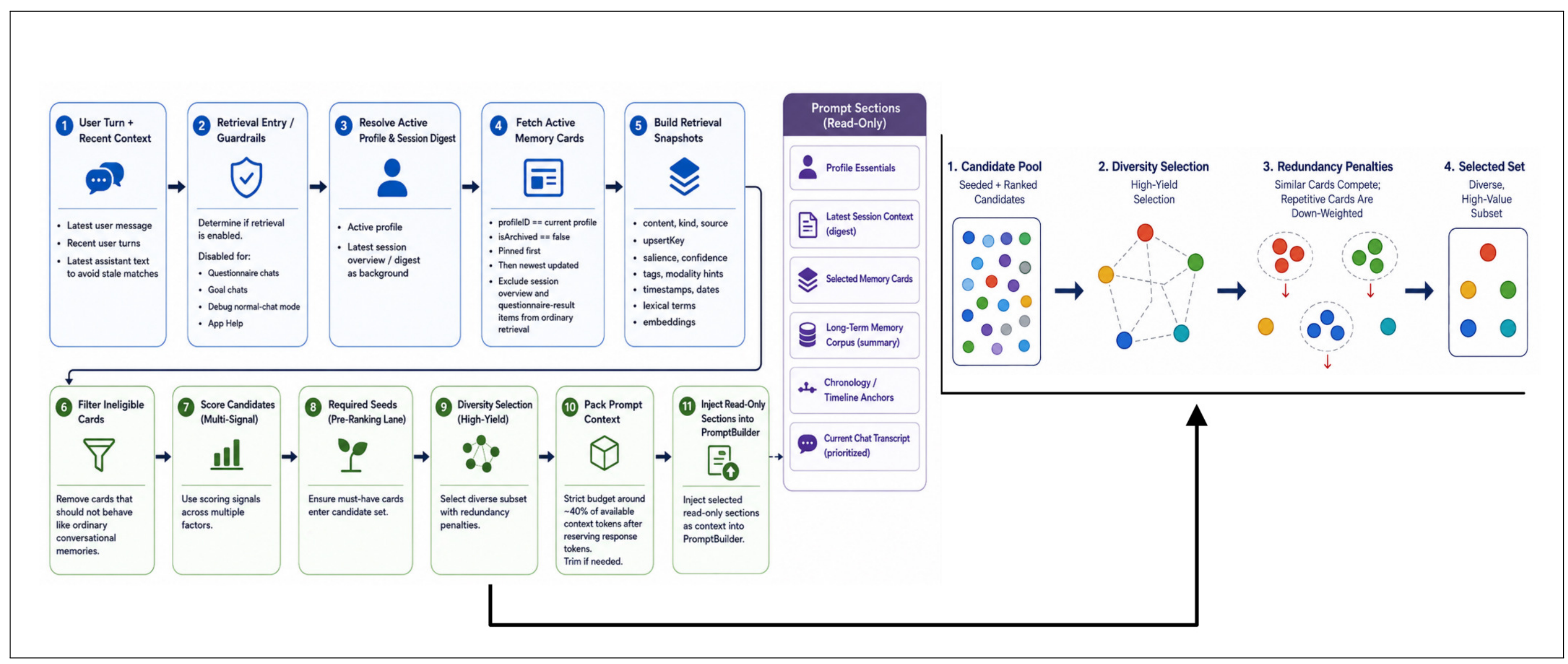

Fig. 2. Memory Notebook view showing user-inspectable structured memory cards.

## 6. Evaluation of Runtime Performance

Runtime performance for all three model options is evaluated under realistic on-device conversation conditions. The primary metrics are time to first token, tokens per second, and context-length sensitivity (Liang et al., 2023; Xu et al., 2024).

Time to first token measures the delay between submitting the model request and receiving the first streamed token, either as the model's answer for the small and medium Gemma models or as the first chain-of-thought token for the largest Gemma 4 model. Tokens per second measures generation speed after the first token. Context-length sensitivity measures how these metrics degrade as the maximum number of memory cards grows.

This evaluation was performed on an Apple iPhone Air, 256 GB, with Low Power Mode and the Adaptive Power battery setting disabled. Runs were performed with battery charge above 50%, and sufficient time between model evaluations was provided to allow the SoC to cool and prevent thermal throttling. Output length was set to verbose, the memory card retrieval maximum was set to 3, and the profile had 85 active memory cards.

The script of prompts was:

1. “I’ve been having trouble sleeping”
2. “I think school stress is mostly responsible”
3. “It’s mostly just wanting to make all my work as perfect as possible”
4. “Can you give me 5 things to try”

## 7. Results

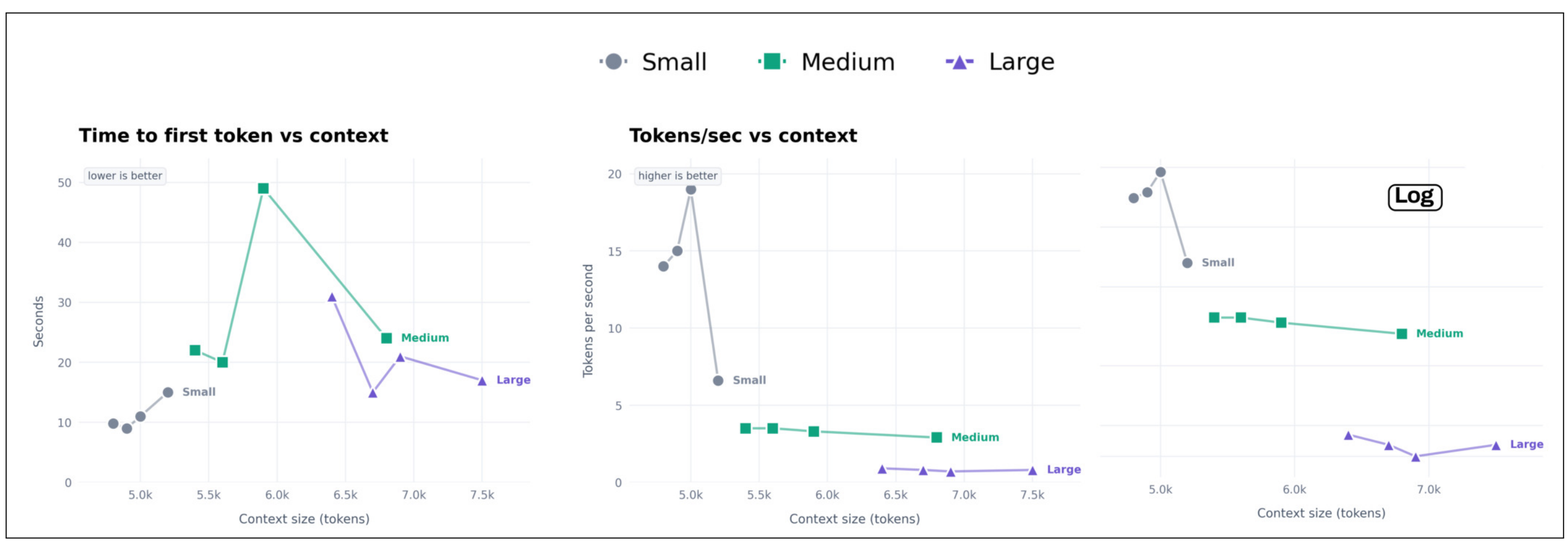

Fig. 3. Runtime performance results for the evaluated local model options under the scripted conversation workload.

## 8. Conclusion

The development and deployment of Psych LM firmly validate the central thesis: for emotionally oriented language model applications, the surrounding architecture is paramount (Boit & Patil, 2025; Lawrence et al., 2024; Packer et al., 2023). The system is not a chat interface wrapped around an LLM; it is a locally-running, memory-augmented, pipeline-controlled construct where the generative model is merely one component in a carefully orchestrated workflow.

The local-first design successfully positioned privacy as a structural property, aligning it with cost control and access. This required treating the entire application as a resource-managed pipeline, overcoming significant on-device constraints such as thermal throttling and unexpected UI-related performance bottlenecks. Crucially, the combination of local inference and the memory corpus, consisting of user-inspectable, validated memory cards, allowed Psych LM to achieve the practical effect of persistent context and longitudinal personalization without requiring a large cloud-based model (Packer et al., 2023; Park et al., 2023; Xu et al., 2024).

The R&D process highlighted that product viability depends on reliability, not abstract capability or model scores on benchmarks (Liang et al., 2023; Liu et al., 2024). Lessons were learned in ensuring behavioral stability: the need for a deterministic orchestration layer to govern model output, the importance of treating prompt payload assembly as a critical resource-allocation problem, and the value of implementing app-mediated validation to mitigate the unreliability of raw memory generation. Ultimately, this work suggests that while LLMs provide the necessary capacity for behavioral coaching, their integration into a stable, usable, and safe application demands intense focus on runtime management, controlled state persistence, and continuous attention to user experience elements such as continuity and UI polish to promote continued use.

Psych LM represents a successful and practical implementation pathway for on-device psychological support systems. It shifts the design discussion away from model size and toward systematic architectural control (Boit & Patil, 2025; Lewis et al., 2020; Packer et al., 2023), proving that complex, context-aware interaction can be reliably achieved under the strict constraints of a local-first mobile environment.

## 9. References

Alizadeh, K., Mirzadeh, I., Belenko, D., Khatamifard, K., Cho, M., Del Mundo, C. C., Rastegari, M., & Farajtabar, M. (2023). LLM in a flash: Efficient large language model inference with limited memory. *arXiv*. https://arxiv.org/abs/2312.11514

Boit, S., & Patil, R. (2025). A prompt engineering framework for large language model–based mental health chatbots: Conceptual framework. *JMIR Mental Health, 12*, e75078. https://doi.org/10.2196/75078

Douze, M., Guzhva, A., Deng, C., Johnson, J., Szilvasy, G., Mazaré, P.-E., Lomeli, M., Hosseini, L., & Jégou, H. (2024). The Faiss library. *arXiv*. https://arxiv.org/abs/2401.08281

Iwaya, L. H., Babar, M. A., Rashid, A., & Wijayarathna, C. (2023). On the privacy of mental health apps: An empirical investigation and its implications for app development. *Empirical Software Engineering, 28*, Article 2. https://doi.org/10.1007/s10664-022-10236-0

Karpukhin, V., Oğuz, B., Min, S., Lewis, P., Wu, L., Edunov, S., Chen, D., & Yih, W.-t. (2020). Dense passage retrieval for open-domain question answering. In *Proceedings of the 2020 Conference on Empirical Methods in Natural Language Processing (EMNLP)* (pp. 6769–6781). Association for Computational Linguistics. https://doi.org/10.18653/v1/2020.emnlp-main.550

Kleppmann, M., Wiggins, A., van Hardenberg, P., & McGranaghan, M. (2019). Local-first software: You own your data, in spite of the cloud. In *Proceedings of the 2019 ACM SIGPLAN International Symposium on New Ideas, New Paradigms, and Reflections on Programming and Software* (pp. 154–178). Association for Computing Machinery. https://doi.org/10.1145/3359591.3359737

Lawrence, H. R., Schneider, R. A., Rubin, S. B., Matarić, M. J., McDuff, D. J., & Jones Bell, M. (2024). The opportunities and risks of large language models in mental health. *JMIR Mental Health, 11*, e59479. https://doi.org/10.2196/59479

Lewis, P., Perez, E., Piktus, A., Petroni, F., Karpukhin, V., Goyal, N., Küttler, H., Lewis, M., Yih, W.-t., Rocktäschel, T., Riedel, S., & Kiela, D. (2020). Retrieval-augmented generation for knowledge-intensive NLP tasks. *Advances in Neural Information Processing Systems, 33*, 9459–9474.

Li, H., Zhang, R., Lee, Y.-C., Kraut, R. E., & Mohr, D. C. (2023). Systematic review and meta-analysis of AI-based conversational agents for promoting mental health and well-being. *npj Digital Medicine, 6*, Article 236. https://doi.org/10.1038/s41746-023-00979-5

Liang, P., Bommasani, R., Lee, T., Tsipras, D., Soylu, D., Yasunaga, M., Zhang, Y., Narayanan, D., Wu, Y., Kumar, A., Newman, B., Yuan, B., Yan, B., Zhang, C., Cosgrove, C., Manning, C. D., Ré, C., Acosta-Navas, D., Hudson, D. A., ... Koreeda, Y. (2023). Holistic evaluation of language models. *Transactions on Machine Learning Research*.

Liu, Z., Zhao, C., Iandola, F., Lai, C., Tian, Y., Fedorov, I., Xiong, Y., Chang, E., Shi, Y., Krishnamoorthi, R., Lai, L., & Chandra, V. (2024). MobileLLM: Optimizing sub-billion parameter language models for on-device use cases. *arXiv*. https://arxiv.org/abs/2402.14905

Packer, C., Wooders, S., Lin, K., Fang, V., Patil, S. G., Stoica, I., & Gonzalez, J. E. (2023). MemGPT: Towards LLMs as operating systems. *arXiv*. https://arxiv.org/abs/2310.08560

Park, J. S., O'Brien, J. C., Cai, C. J., Morris, M. R., Liang, P., & Bernstein, M. S. (2023). Generative agents: Interactive simulacra of human behavior. In *Proceedings of the 36th Annual ACM Symposium on User Interface Software and Technology*. Association for Computing Machinery. https://doi.org/10.1145/3586183.3606763

Reimers, N., & Gurevych, I. (2019). Sentence-BERT: Sentence embeddings using Siamese BERT-networks. In *Proceedings of the 2019 Conference on Empirical Methods in Natural Language Processing and the 9th International Joint Conference on Natural Language Processing* (pp. 3982–3992). Association for Computational Linguistics. https://doi.org/10.18653/v1/D19-1410

Sahoo, P., Singh, A. K., Saha, S., Jain, V., Mondal, S., & Chadha, A. (2024). A systematic survey of prompt engineering in large language models: Techniques and applications. *arXiv*. https://arxiv.org/abs/2402.07927

White, J., Fu, Q., Hays, S., Sandborn, M., Olea, C., Gilbert, H., Elnashar, A., Spencer-Smith, J., & Schmidt, D. C. (2023). A prompt pattern catalog to enhance prompt engineering with ChatGPT. *arXiv*. https://arxiv.org/abs/2302.11382

Xu, J., Li, Z., Chen, W., Wang, Q., Gao, X., Cai, Q., & Ling, Z. (2024). On-device language models: A comprehensive review. *arXiv*. https://arxiv.org/abs/2409.00088